\documentclass[sigconf, nonacm]{acmart}

\AtBeginDocument{
  \providecommand\BibTeX{{
    \normalfont B\kern-0.5em{\scshape i\kern-0.25em b}\kern-0.8em\TeX}}}

\usepackage[utf8]{inputenc} 
\usepackage[T1]{fontenc}    
\usepackage{hyperref}       
\usepackage{url}            
\usepackage{booktabs}       
\usepackage{amsfonts}       
\usepackage{nicefrac}       
\usepackage{microtype}      
\usepackage{xcolor}        
\usepackage{graphicx}
\usepackage{subcaption} 
\usepackage[ruled,vlined]{algorithm2e} 
\usepackage{multirow}
\usepackage{booktabs}
\usepackage{enumitem}
\usepackage{mathtools}

\usepackage{amsthm}
\usepackage{color}

\newcommand{\cut}[1]{{}}

\newcommand{\tA}{{\tilde{\vA}}}
\newcommand{\tD}{{\tilde{\vD}}}
\newcommand{\tL}{{\tilde{\vL}}}
\newcommand{\hA}{{\hat{\vA}}}
\newcommand{\hD}{{\hat{\vD}}}

\newcommand{\vA}{{\mathbf{A}}}

\newcommand{\vD}{{\mathbf{D}}}

\newcommand{\vI}{{\mathbf{I}}}

\newcommand{\vL}{{\mathbf{L}}}
\newcommand{\vM}{{\mathbf{M}}}

\newcommand{\vU}{{\mathbf{U}}}

\newcommand{\vX}{{\mathbf{X}}}
\newcommand{\vY}{{\mathbf{Y}}}

\newcommand{\cG}{{\mathcal{G}}}

\newcommand{\cV}{{\mathcal{V}}}

\newcommand{\RR}{\mathbb{R}}

\makeatletter
\let\@@span\span
\def\sp@n{\@@span\omit\advance\@multicnt\m@ne}
\makeatother

\newcommand{\bc}{\begin{center}}
\newcommand{\ec}{\end{center}}

\newcommand{\bdm}{\begin{displaymath}}
\newcommand{\edm}{\end{displaymath}}

\newcommand{\beq}{\begin{equation}}
\newcommand{\eeq}{\end{equation}}

\newcommand{\bfl}{\begin{flushleft}}
\newcommand{\efl}{\end{flushleft}}

\newcommand{\bt}{\begin{tabbing}}
\newcommand{\et}{\end{tabbing}}

\newcommand{\beqn}{\begin{align}}
\newcommand{\eeqn}{\end{align}}

\newcommand{\beqs}{\begin{align*}} 
\newcommand{\eeqs}{\end{align*}}  

\usepackage{array}

\title{Automated Polynomial Filter Learning \\ for Graph Neural Networks}

\author{Wendi Yu}

\orcid{0009-0005-4641-7914}
\affiliation{
  \institution{Tongji University}
  \city{Shanghai}
  \country{China}
}
\email{yuwendi@tongji.edu.cn}

\author{Zhichao Hou}

\affiliation{
  \institution{North Carolina State University}
  \streetaddress{1 Th{\o}rv{\"a}ld Circle}
  \city{Raleigh}
  \state{North Carolina}
  \country{USA}
  }
\email{zchou0807@gmail.com}

\author{Xiaorui Liu}

\affiliation{
  \institution{North Carolina State University}
  \city{Raleigh}
  \state{North Carolina}
  \country{USA}
}
\email{xliu96@ncsu.edu}



\begin{document}

\begin{abstract}

Polynomial graph filters have been widely used as guiding principles in the design of Graph Neural Networks (GNNs). Recently, the adaptive learning of the polynomial graph filters has demonstrated promising performance for modeling graph signals on both homophilic and heterophilic graphs, owning to their flexibility and expressiveness. 
In this work, we conduct a novel preliminary study to explore the potential and limitations of polynomial graph filter learning approaches, revealing a severe overfitting issue. 
To improve the effectiveness of polynomial graph filters, we propose Auto-Polynomial, a novel and general automated polynomial graph filter learning framework that efficiently learns better filters capable of adapting to various complex graph signals. Comprehensive experiments and ablation studies demonstrate significant and consistent performance improvements on both homophilic and heterophilic graphs across multiple learning settings considering various labeling ratios, which unleashes the potential of polynomial filter learning.
\end{abstract}

\maketitle

\section{Introduction}

In recent years, Graph neural networks (GNNs) have received great attention due to their strong prediction performance in various graph learning tasks, such as node classification~\cite{oono2019graph}, link prediction~\cite{zhang2018link} and graph classification~\cite{errica2019fair}.
They have been successfully applied to numerous real-world problems since the structured data in many real-world scenarios, such as social networks~\cite{wu2020graph}, recommendation systems~\cite{wu2022graph} and traffic networks~\cite{jiang2022graph}, can be ubiquitously represented as graphs. GNNs' superiority is due to their capability in capturing and incorporating both node feature and graph structure information through message passing or feature aggregation operations that can be inspired from either spatial domain or spectral domain~\cite{grl_william, ma2020deep}. 

While numerous GNN architecture designs exist, recent studies suggest that many popular GNNs can be understood as operating various polynomial graph spectral filtering on the graph signal~\cite{Chebnet,kipf2016semi, chien2021GPR-GNN, levie2018cayleynets, xu2018graphWavelet, bianchi2021ARMA}. From the polynomial graph filter point of view, these GNNs can be broadly categorized into two classes, depending on whether the polynomial graph filters can be either manually designed (e.g., GCN\cite{kipf2016semi}, APPNP~\cite{appnp}, SGC~\cite{wu2019simplifying}, GNN-LF/HF~\cite{zhu2021interpreting}) or learned from graph data (e.g., ChebNet~\cite{Chebnet}, GPRGNN~\cite{chien2021GPR-GNN}, BernNet~\cite{he2021bernnet}). 
For instance, GCN adopts a simplified first-order Chebyshev polynomial, and APPNP uses Personalized PageRank to design the graph filter. On the contrary, ChebNet, GPRGNN, and BernNet choose to approximate the graph filter with various polynomials using the Chebyshev basis, Monomial basis, and Bernstein basis respectively whose polynomial coefficients are trainable parameters that can be learned from data.

A large body of literature has demonstrated that in practice, manually designed graph filters perform pretty well on homophilic graphs where nodes with similar features or class labels tend to be connected with each other~\cite{kipf2016semi, wu2019sgc, appnp, hu2020open}.
In these cases, simple low-frequency filters, such as those filters in GCN~\cite{kipf2016semi}, SGC~\cite{wu2019sgc}, and APPNP~\cite{appnp}, typically model the homophilic graph signal well since the graph signal is supposed to be smooth over the edges of the graph. However, it is much more challenging to model the graph signal for heterophilic graphs where the graph signal is not necessarily smooth and highly dissimilar nodes tend to form many edges~\cite{ma2022is, zhu2020beyond, zhu2021graph}. For example, in dating networks, people with opposite gender tend to contact each other~\cite{zhao2013user}, and different types of amino acids are more likely to link together to form protein structures~\cite{zhu2020beyond}, etc. 
Therefore, the low-pass filter via a simple neighborhood aggregation mechanism in GNNs is problematic and not well-suited for heterophilic graphs. 
In fact, the proper prior knowledge of the heterophilic graph signal is generally unknown. In these cases, learnable polynomial graph filters, such as those in ChebNet, GPRGNN, and BernNet, exhibit some encouraging performance on many heterophilic datasets since they are able to customize graph filters for graphs in order to accommodate their respective assumptions and adapt to various types of graphs by learning the coefficients of polynomial filters.
This indicates the profound potential of graph polynomial filter learning.

While GNNs with polynomial graph filter learning are promising due to their great flexibility and powerful expressive capacity, we discover that they still face various limitations through our preliminary study in Section~\ref{sec:pre}. First, we reveal that the learning of polynomial filters is quite sensitive to the initialization of polynomial filters, which leads to unstable performance and a significantly large variance. 
Second, we find that they sometimes can exhibit poor performance in practice, especially under semi-supervised settings. 
Third, their performance often underperforms simple graph-agnostic multi-layer perceptrons (MLPs) on heterophilic graphs, which indicates the failure in graph signal modeling. However, the reasons behind these failures are still obscure.

In this work, we aim to explore the potential and limitations of the widely used polynomial graph filter learning technique for modeling complex graph signals in GNNs. Our contribution to be summarized as the following:
\begin{itemize}[leftmargin=0.2in]
\item We conduct a novel investigation to study whether existing approaches have fully realized the potential of graph polynomial filter learning. Surprisingly, we discover that graph polynomial filter learning suffers from a severe overfitting problem, which provides a potential explanation for its failures. 
\item To overcome the overfitting issue, we propose Auto-Polynomial, a novel and general automated polynomial filter learning framework to enhance the performance and generalization of polynomial graph spectral filters in GNNs. 
\item Comprehensive experiments and ablation studies demonstrate significant and consistent performance improvements on both homophilic and heterophilic graphs across multiple learning settings, which efficiently unleashes the potential of polynomial filter learning.
\end{itemize}

\section{Preliminary}
\label{sec:pre}
\textbf{Notations.}
An undirected graph can be represented by $\mathcal{G} =(\mathcal{V}, \mathcal{E})$ where $\mathcal{V} =\left \{ \upsilon _{1},...,\upsilon  _{n}  \right \}$ is the set of the  nodes and $\mathcal{E} =\left \{  e_{1},...,e_{m}  \right \} $ is the set of the edges. Suppose that the nodes are associated with the node feature matrix $\vX \in \mathbb{R}^{n\times d}$, where $n$ denotes the number of
nodes and $d$ denotes the number of features per node. 
The graph structure of $\mathcal{G}$ can be represented by the adjacency matrix $\vA$, and $\hA=\vA+\vI$ denotes the adjacency matrix with added self-loops. 
Then the symmetrically normalized adjacency matrix and graph Laplacian matrix can be defined as
$\tA=\tD^{-\frac{1}{2}} \hA \tD^{-\frac{1}{2}}$ and $\tL=\vI-\tA = \mathbf{I}-\hD^{-\frac{1}{2}} \hA \hD^{-\frac{1}{2}}$ respectively, where $\hD$ is the diagonal degree matrix of $\hA$. Next, we will briefly introduce the polynomial graph filter as well as the concepts of homophilic and heterophilic graphs.

\subsection{Polynomial Graph Filter}
\label{sec:poly-filter}
Spectral-based GNNs operate graph
convolutions in the spectral domain of the graph Laplacian. 
Many of these methods utilize
polynomial graph spectral filters to achieve this operation. The polynomial graph signal filter can be formulated as follows:
\vspace{-0.1in}
\begin{align}
    \vY = \vU diag [h(\lambda_1), \dots, h(\lambda_n)] \vU^\top \vX = \vU h(\mathbf{\Lambda}) \vU^\top \vX \approx \sum_{k=1}^K \theta_k \tL^k \vX, \nonumber
\end{align}
where $\vX\in\RR^{n\times d}$ is the input signal and $\vY\in\RR^{n\times d}$ is the output signal after filtering. $\tL=\vU \mathbf{\Lambda} \vU^\top$ denotes the eigendecomposition of the symmetric normalized Laplacian matrix $\tL$, 
where $\vU$ denotes the matrix of eigenvectors and $\mathbf{\Lambda}=diag[\lambda_1,\dots,\lambda_n]$ is the diagonal matrix of eigenvalues. $h(\lambda)$ is an arbitrary filter function that can be approximated by various 
polynomials $h_\Theta(\lambda) = \sum_{k=0}^K \theta_k \lambda^k, \lambda \in [0,2]$, where the polynomial coefficients $\Theta=\{\theta_k \}_{k=0}^K$
determine the shape and properties of the $k$-th order filter. 

The filter approximation using polynomials has multiple unique advantages. First, the filtering operation represented by polynomial filters can be equivalently and efficiently computed in the spatial domain using simple recursive feature aggregations, i.e., $\sum_{k=1}^K \theta_k \tL^k \vX$, without resorting to expensive eigendecomposition that is infeasible for large-scale graphs. 
Second, the polynomial filters enable localized computations such that it only requires $k$-hop feature aggregation, and only $k$ parameters are needed compared with the non-parametric filter $h(\mathbf{\Lambda}) = diag[h(\lambda_1), \dots, h(\lambda_n)]$. Most importantly, the polynomial filter is flexible and general to represent filters with various properties~\cite{he2021bernnet} such as low-pass filters, high-pass filters, band-pass filters, band-rejection filters, comb filters, etc.
Generally, the polynomial graph filter can be written in the general form:
\begin{equation}
\label{eq:poly-filter}
    \vU h(\mathbf{\Lambda}) \vU^\top \vX \approx \vU h_\Theta(\mathbf{\Lambda}) \vU^\top = \sum_{k=0}^{K}\theta  _{k}P_k(\vM),
\end{equation}
where $\{\theta_k \}_{k=0}^K$ are the polynomial coefficients, $\{P_k(x)\}_{k=0}^K$ are the polynomial basis, and $\vM$ is the basic matrix (e.g., normalized graph Laplacian matrix $\tL$ or normalized adjacency matrix $\tA$).
Among all the works based on polynomial filters, we will introduce the foundational ChebNet~\cite{kipf2016semi} and other improved models such as GPRGNN~\cite{chien2020adaptive} and BernNet~\cite{he2021bernnet}.

\vspace{0.05in}
\noindent \textbf{ChebNet: Chebyshev polynomials.}
ChebNet is a foundational work that uses the Chebyshev polynomials to approximate the graph spectral filter:
\begin{equation}
\label{eq:chebnet}
    \sum_{k=0}^{K}\theta _{k}T_{k}
    (\frac{2\tL}{\lambda_{max}} - \vI),  
\end{equation}
  
\noindent where 
$\frac{2\tL}{\lambda_{max}} - \vI$ 
denotes the rescaled Laplacian matrix, and $\lambda_{max}$ is the largest eigenvalue of $\tL$.
The Chebyshev polynomial $T_{k}(x)$ of order $k$ can be computed by the stable recurrence relation $T_{k}(x) =2xT_{k-1}(x)-T_{k-2}(x)$, with $T_{0}(x) =1$ and $T_{k}(x)=x$. 
The polynomial filter Eq.~\eqref{eq:chebnet} can be regarded as a special case of the general form in Eq.~\eqref{eq:poly-filter} when we set $P_k(x)=T_k(x)$ and 
$\vM=\frac{2\tL}{\lambda_{max}} - \vI$.

\vspace{0.05in}
\noindent \textbf{GPRGNN: Monomial polynomials.}
GPRGNN approximates spectral graph convolutions by Generalized PageRank~\cite{chien2021GPR-GNN}. The spectral polynomial filter of GPRGNN can be represented as follows:
\begin{equation}
\label{eq:GPRGNN}
    \sum_{k=0}^{K}\theta_{k} \tA^{k},
\end{equation}
which is equivalent to the general form in Eq.~\eqref{eq:poly-filter} when we set $P_k(x)=x^k$ and $\vM=\tA$.

\vspace{0.05in}
\noindent \textbf{BernNet: Bernstein polynomials.}
The principle of BernNet is to approximate any filter on the normalized Laplacian spectrum of a graph using K-order Bernstein polynomials. By learning the coefficients of the Bernstein Basis, it can adapt to different signals and get various spectral filters.
The spectral filter of BernNet can be formulated as:
\begin{equation}
\label{eq:bernnet}
    \sum_{k=0}^{K} \theta_{k} \frac{1}{2^{K}}\left(\begin{array}{l}
K \\
k
\end{array}\right)(2\vI-\tL)^{K-k} \tL^{k},
\end{equation}
which is equivalent to the general form in Eq.~\eqref{eq:poly-filter} when we set $P_k(x)=\frac{1}{2^{K}}\left(\begin{array}{l}K \\k\end{array}\right)(2-x)^{K-k} x^{k}$ and $\vM=\tL$.

In the aforementioned GNNs, the coefficients, i.e., $\Theta$, determining the polynomial filters are treated as learnable parameters together with other model parameters that are learned by gradient descent algorithms based on the training loss. They have shown encouraging performance on both homophilic and heterophilic graphs since the filter can be adapted to model various types of graph signals whose proper assumptions are generally complex and unknown.

\subsection{Homophily and Heterophily}
Graphs can be classified into homophilic graphs and heterophilic graphs based on the concept of node homophily~\cite{zhu2020beyond, pei2020geom, ma2022is}. In homophilic graphs, nodes with similar features or belonging to the same class tend to connect with each other. For example, in citation networks, papers in the same research field tend to cite each other~\cite{ciotti2016homophily}. In other words, homophilic graphs usually satisfy the smoothness assumption that the graph signal is smooth over the edges. 
The majority of GNNs are under homophily assumption, and their manually designed low-pass filters filter out the high-frequency graph signal via simple neighborhood aggregation mechanisms that propagate the graph information. For instance, GCN~\cite{kipf2016semi} uses the first-order Chebyshev polynomial as graph convolution, which is proven to be a fixed low-pass filter~\cite{nt2019revisiting}. APPNP~\cite{appnp} uses a Personalized Pagerank(PPR) to set fixed filter coefficients, which has been shown to surpass the high-frequency graph signals~\cite{chien2021GPR-GNN}. Generally, these manually designed graph filters exhibit superior performance in various prediction tasks on homophilic graphs.

On the other hand, in heterophilic graphs, nodes with different features or belonging to different classes tend to connect with each other. For example, in dating networks, most users tend to establish connections with individuals of the opposite gender, exhibiting heterophily~\cite{zhu2021graph}. In molecular networks, protein structures are more likely to be composed of connections between different types of amino acids~\cite{zhu2020beyond}. In order to quantify the degree of homophily of the graphs, Pei et al~\cite{pei2020geom} have proposed a metric to measure the level of node homophily in graphs:
\begin{align}
H(\cG)=\frac{1}{|\mathcal{V}|} \sum_{v \in \mathcal{V}} \frac{\left|\left\{u \in \mathcal{N}(v): y_{v}=y_{u}\right\}\right|}{|\mathcal{N}(v)|}
\end{align}
where $\mathcal{N}(v)$ denotes the neighbour set of node $v$, and $y_{v}$ denotes the label of node $v$. This metric $H(\cG)$ measures the average probability that an edge connects two nodes of the same label over all nodes in $\cV$.
When $H(\cG) \rightarrow 1$, it indicates the graph is of high homophily since the edge always connects nodes with the same label, while $H(\cG) \rightarrow 0$ indicates the graph is of high heterophily where edge always connect nodes with different labels.
In fact, the proper prior knowledge of the heterophilic graphs is generally unknown. Therefore, those classic GNNs cannot perform well on heterophilic graphs since their expressive power is limited by fixed filters.

In recent years, many works have been actively exploring GNNs under heterophily. 
Some examples of heuristic approaches include MixHop~\cite{abu2019mixhop}, H2GCN~\cite{zhu2020generalizing}, and GCNII~\cite{chen2020simple}. 
For example, MixHop~\cite{abu2019mixhop} is an exemplary method that aggregates messages from higher-order neighbors, thereby enabling the integration of information from different distances. Another notable approach is H2GCN~\cite{zhu2020generalizing}, which aims to address heterophily by separating the ego and neighbor embeddings. GCNII~\cite{chen2020simple} adopts various residual connections and more layers to capture distant information in graphs.
They all attempt to aggregate information from higher-order neighbors.
There also exist multiple approaches inspired by polynomial graph filters~\cite{Chebnet, chien2021GPR-GNN, he2021bernnet}. These polynomial graph filter learning approaches have better interpretability than heuristic methods, and they exhibit encouraging performance on some heterophilic graphs,
owing to the flexibility and adaptivity to graph signals of various properties. However, there is no consistent performance advantage for either heuristic or graph polynomial filter learning approaches. Both types of methods sometimes underperform graph-agnostic MLPs.

\section{Overfitting of graph polynomial filter learning}
\label{sec:overfit}
In this section, we design novel preliminary experiments to investigate the potential and limitations of polynomial graph filter learning in GNNs.
\begin{figure*}[!htp]
\vspace{-0.2in}
    \centering
    \begin{subfigure}[b]{0.48\textwidth}
        \centering
        \includegraphics[width=\textwidth]{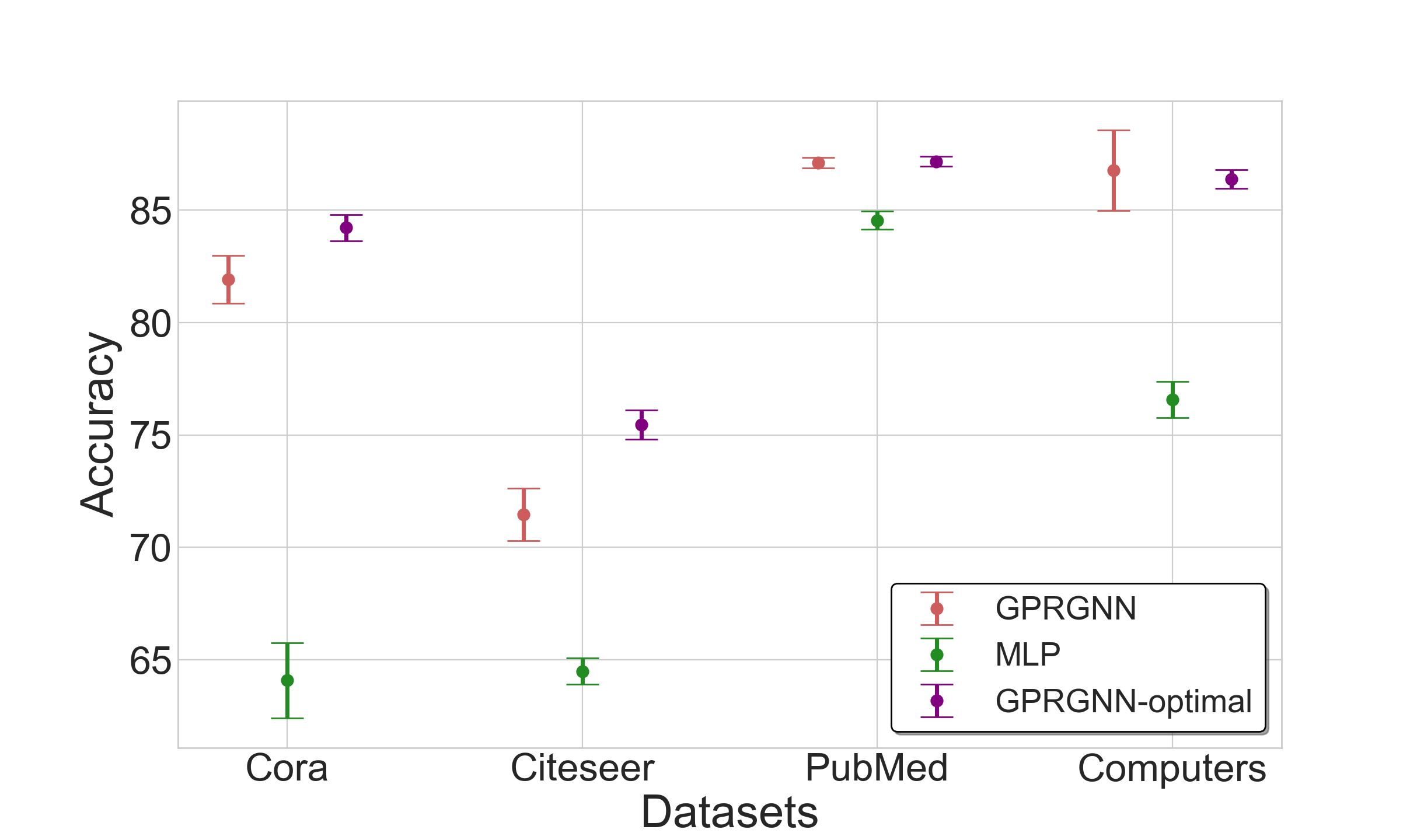}
        \caption{Homophilic datasets.}
        \label{fig:subfig1}
    \end{subfigure}
    \begin{subfigure}[b]{0.48\textwidth}
        \centering
        \includegraphics[width=\textwidth]{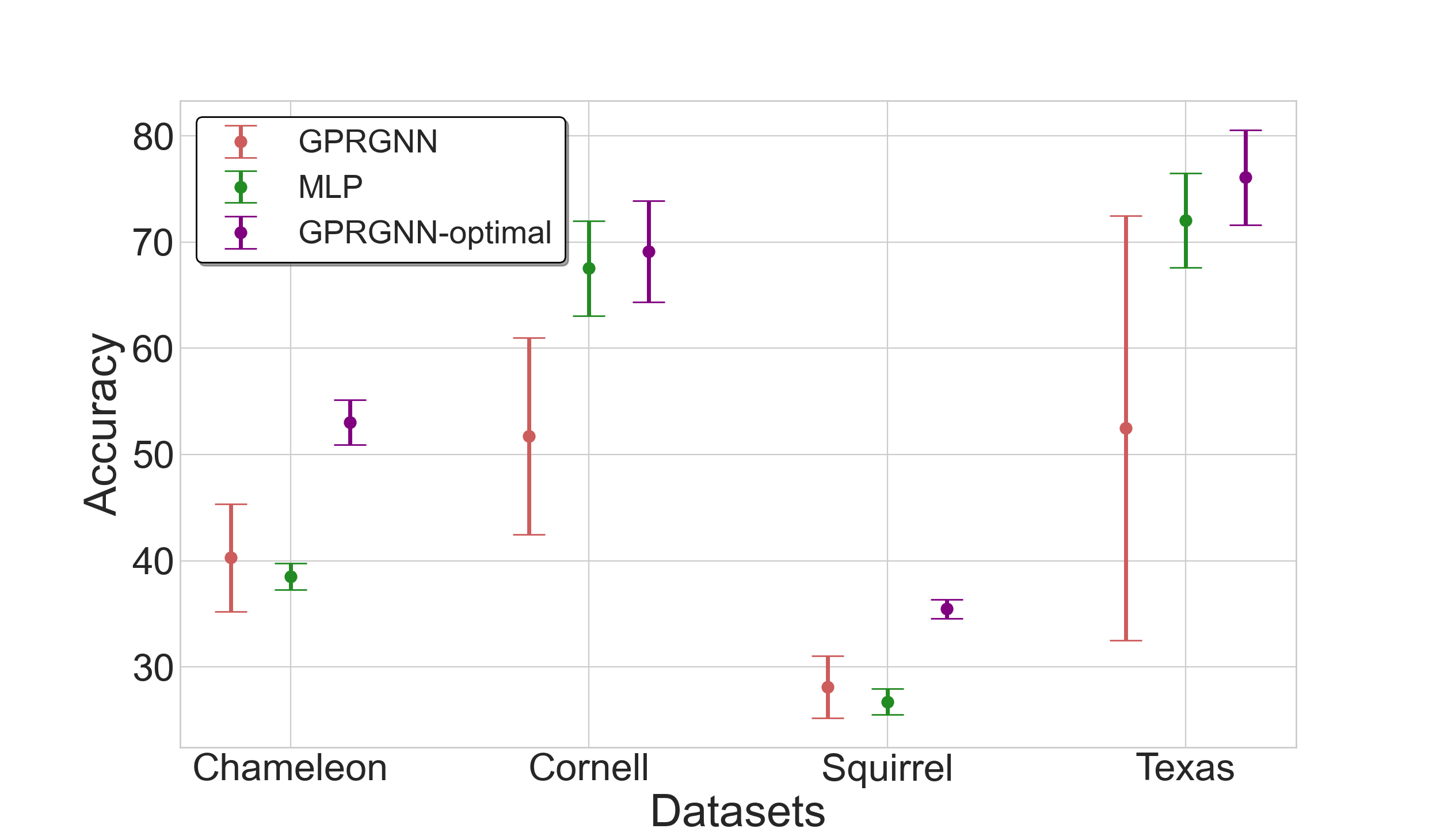}
        \caption{Heterophilic datasets.}
        \label{fig:subfig2}
    \end{subfigure}
    \caption{The comparison between MLP, GPRGNN, and GPRGNN-optimal on homophilic and heterophilic datasets.}
    \label{fig:main}
\end{figure*}

\begin{figure*}[!htp]
  \centering
  \subcaptionbox{Cora\label{fig:subfig1}}{\includegraphics[width=0.26\linewidth]{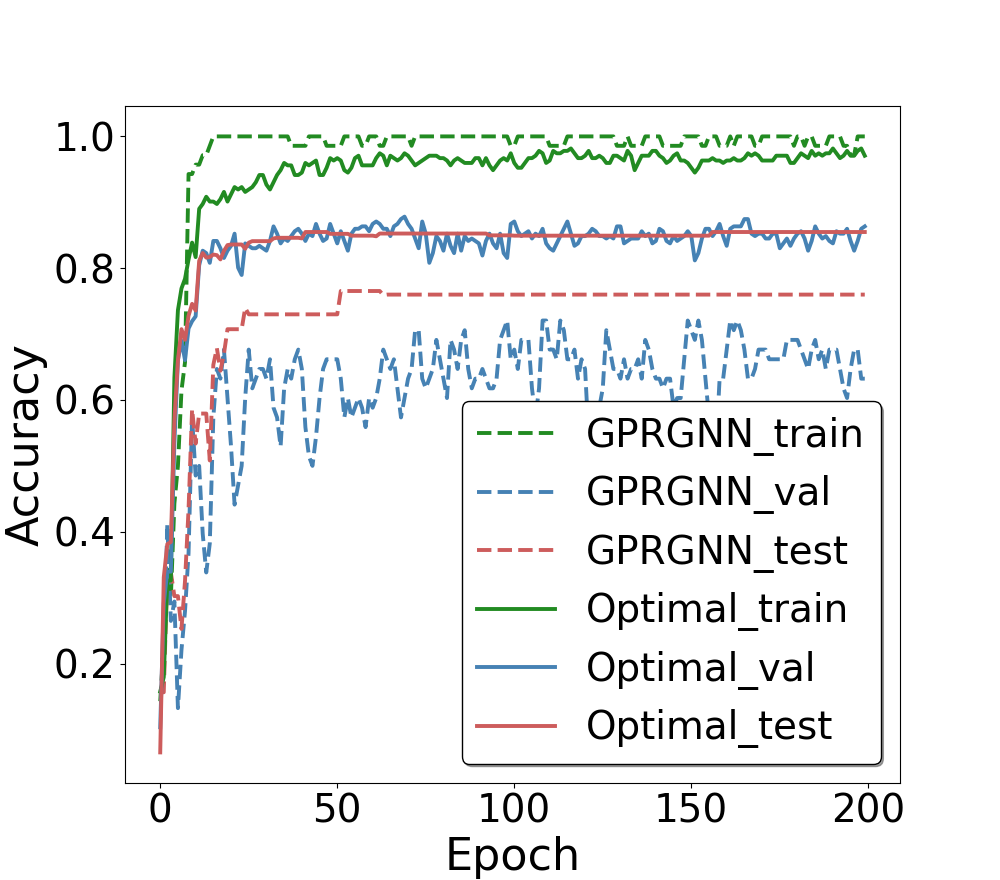}}
  \hspace{-15pt} 
  \subcaptionbox{Citeseer\label{fig:subfig2}}{\includegraphics[width=0.26\linewidth]{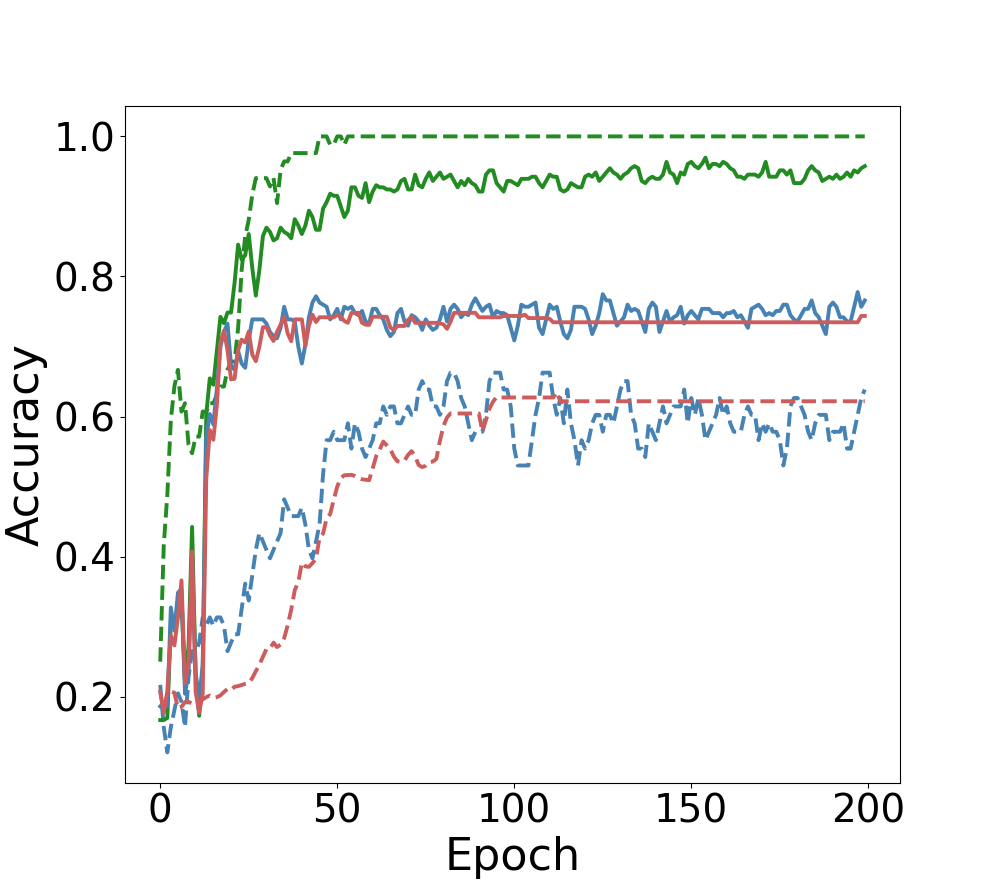}}\hspace{-15pt} 
  \subcaptionbox{PubMed\label{fig:subfig3}}{\includegraphics[width=0.26\linewidth]{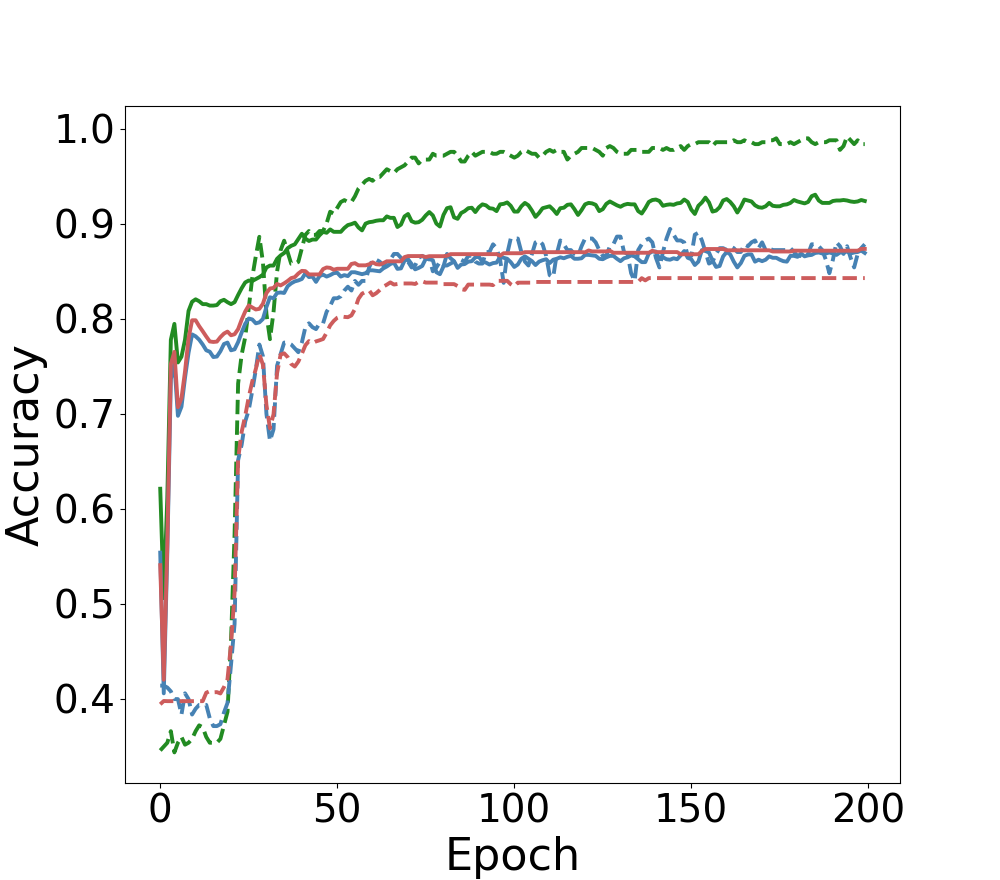}}
  \hspace{-15pt}  
  \subcaptionbox{Computers\label{fig:subfig4}}{\includegraphics[width=0.26\linewidth]{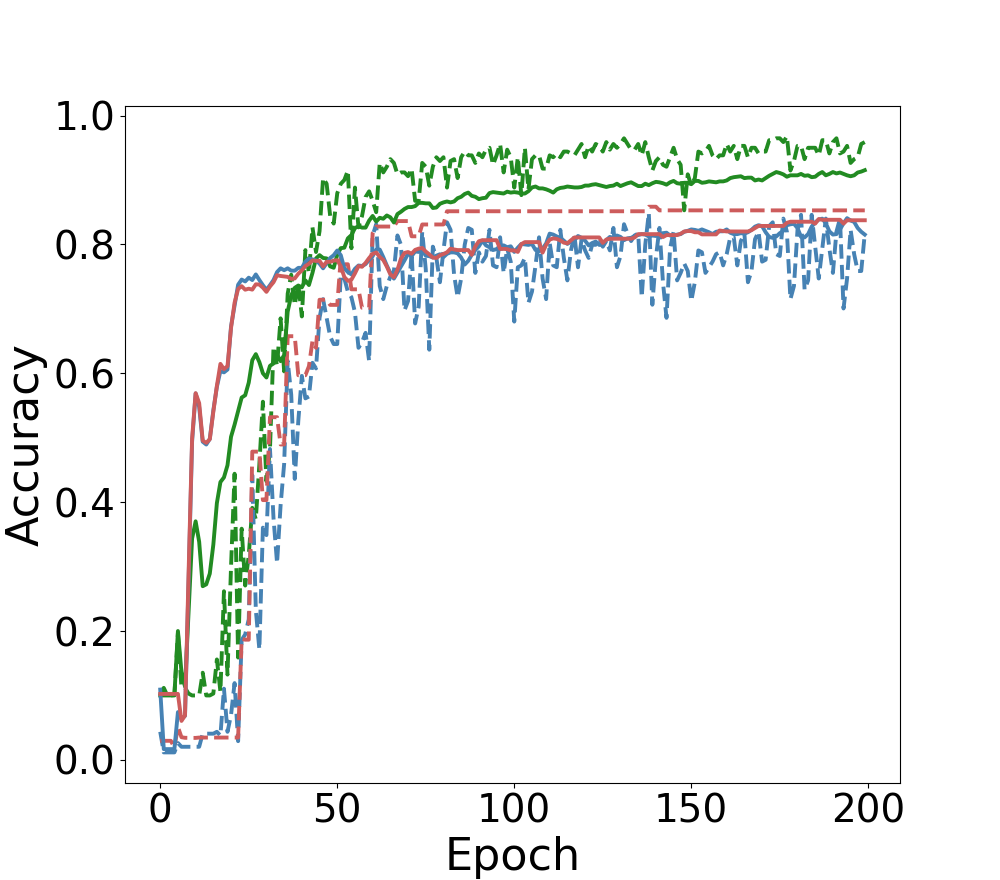}} 
  \\
  \subcaptionbox{Chameleon\label{fig:subfig5}}{\includegraphics[width=0.26\linewidth]{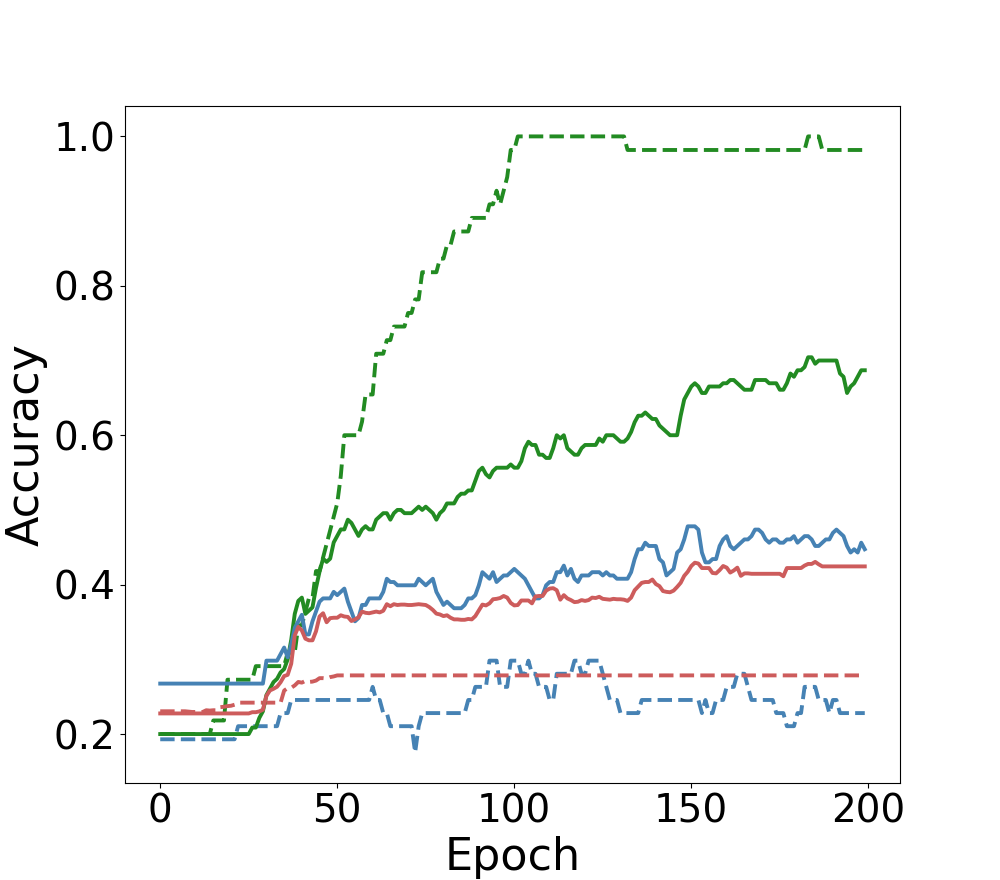}}\hspace{-15pt}
  \subcaptionbox{Cornell\label{fig:subfig6}}{\includegraphics[width=0.26\linewidth]{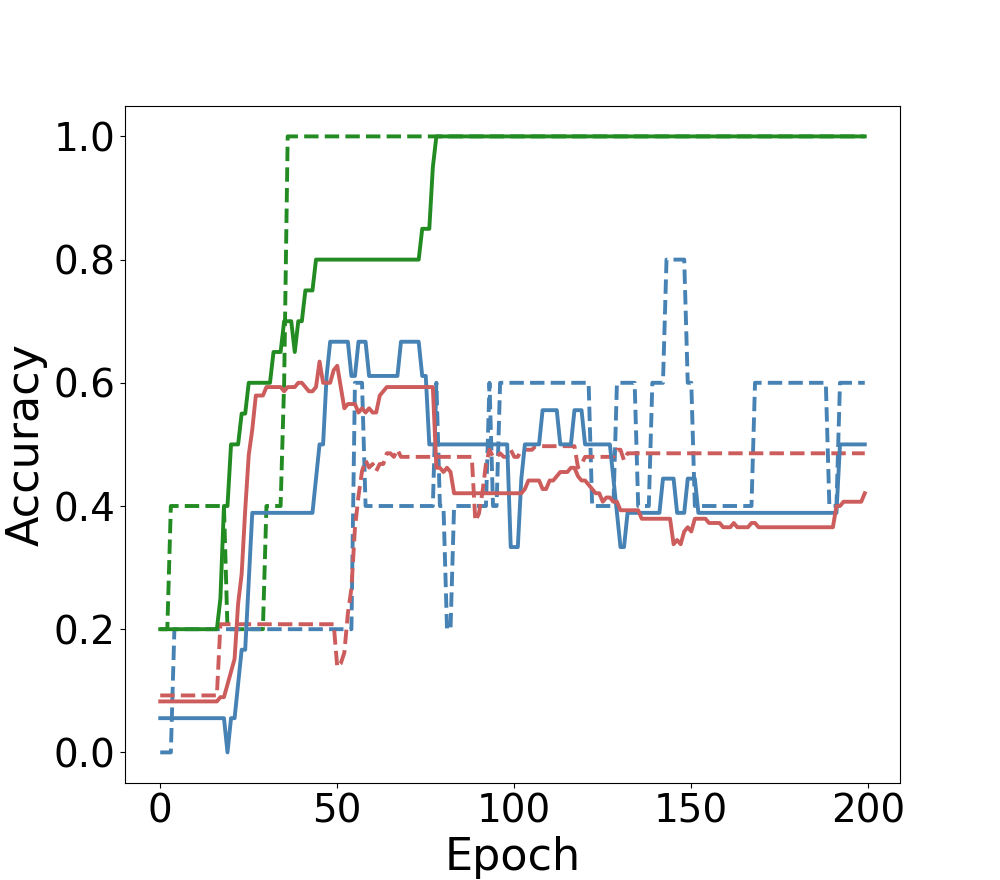}}\hspace{-15pt} 
  \subcaptionbox{Squirrel\label{fig:subfig7}}{\includegraphics[width=0.26\linewidth]{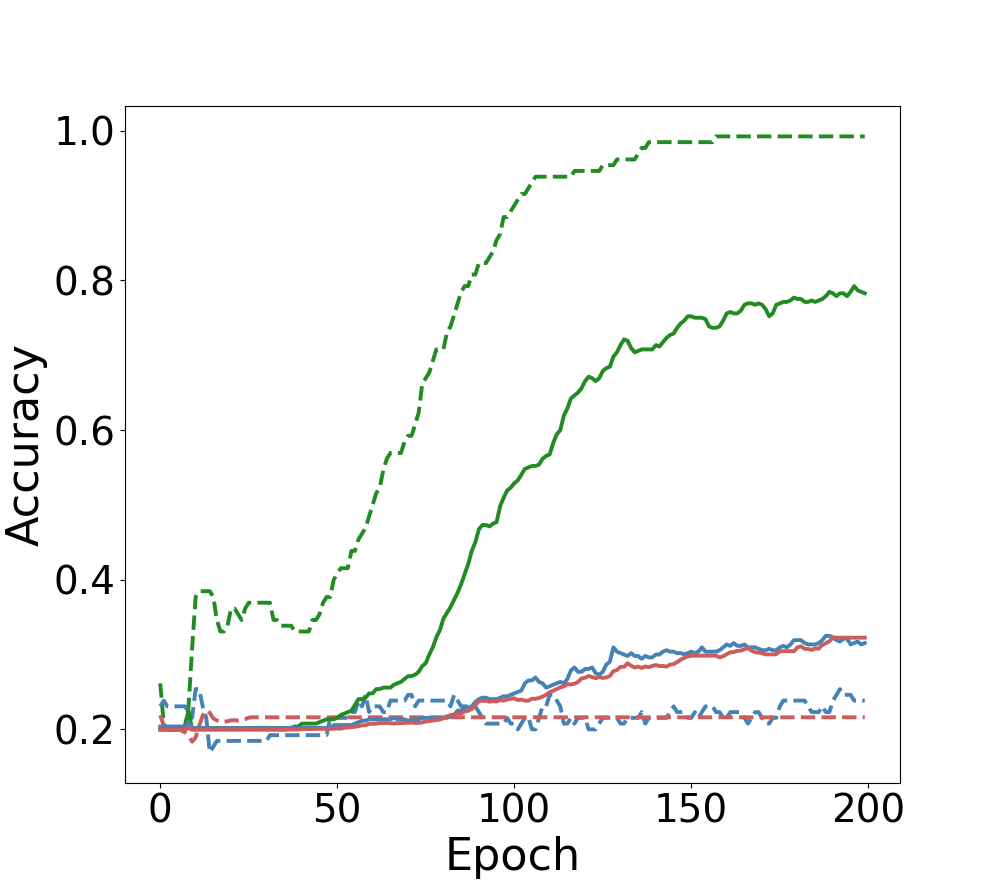}}\hspace{-15pt} 
  \subcaptionbox{Texas\label{fig:subfig8}}{\includegraphics[width=0.26\linewidth]{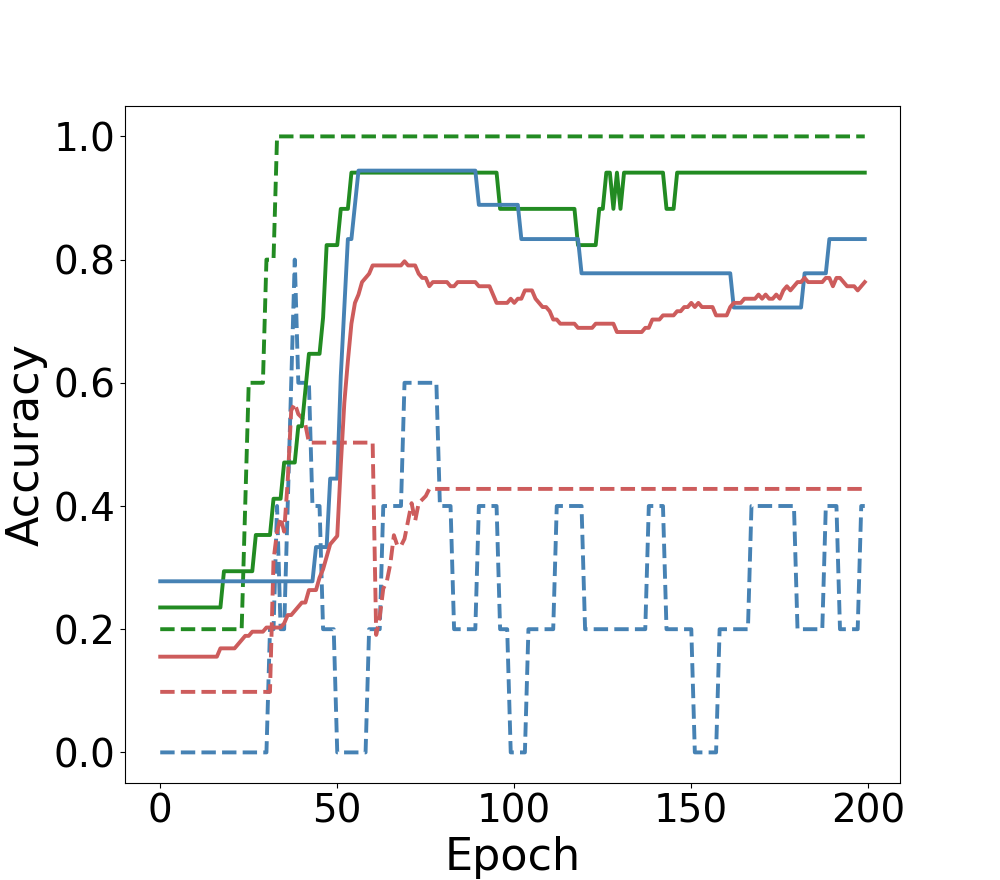}}
  \caption{The training/validation/test accuracy of GPRGNN and GPRGNN-optimal in the training process.}
  \label{fig:figure2}
\end{figure*}

\subsection{Search of optimal polynomials}
\label{sec:search-optimal}
Theoretically speaking, polynomial-based GNNs have strong flexibility, adaptivity, and expressiveness as discussed in Section~\ref{sec:pre}. 
Although they achieve encouraging performance in some cases, they often fail to show consistent improvement and advantages over simpler designs in practice. 
On one hand, polynomial-based GNNs do not exhibit notably better performance than 
manually designed GNNs such as APPNP~\cite{appnp} and GCNII~\cite{chen2020simple} on homophilic datasets even though these manually designed GNNs simply emulate low-pass filters and can be covered and approximated as special cases of polynomial-based GNNs. 
On the other hand, polynomial-based GNNs~\cite{Chebnet, chien2021GPR-GNN, he2021bernnet} 
tend to learn inappropriate polynomial weights and often exhibit inferior performance. For example, ChebNet performs noticeably worse than its simplified version, GCN~\cite{kipf2016semi}, which only utilizes the first two Chebyshev polynomials;
The improved variants such as GPRGNN~\cite{chien2021GPR-GNN} and BernNet~\cite{he2021bernnet} can not consistently outperform heuristically and manually designed alternatives such as H2GCN~\cite{zhu2020generalizing}; and they even underperform graph-agnostic MLPs in some cases.

The failures of polynomial-based GNNs are counterintuitive considering their strong expressively, which then leads to the following critical question:
``\textbf{Have we achieved the full expressiveness and adaptivity of polynomial graph filter learning?}''
To answer this question, we design a novel experiment to explore the full potential of polynomial filters, i.e., the best performance that the optimal polynomial filter can potentially achieve. 
Specifically, we choose the monomial polynomial (i.e., $\sum_{k=0}^{K}\theta_{k} \tA^{k}$) as designed in GPRGNN and perform a brute-force search to explore the optimal polynomial filter by a grid search on the polynomial coefficients $\{\theta_k\}_{k=0}^K$ based on validation accuracy. We denote the GPRGNN with this optimal polynomial filter as GPRGNN-optimal. We compare the performance with the baseline GPRGNN whose polynomial coefficients are learnable parameters guided by training loss. 

In detail, we evaluate the performance on semi-supervised node classification tasks on some classic homophilic and heterophilic datasets. 
In practice, it is computationally expensive to conduct an exhaustive brute-force search on the polynomial coefficients if the polynomial order $K$ is large or the search granularity is fine since the search space will exponentially increase.
To reduce the search space for this preliminary study, we set the polynomial order $K=2$ so that we only need to do a grid search over $3$ parameters in the range $\{-0.9, -0.5, -0.2, -0.1, -0.05, 0, 0.05, 0.1, 0.2, 0.5, 0.9\}$. While this coarse search can not cover the whole search space, the results provide performance lower bounds for the optimal polynomial filter.
For all models, we adopt the sparse splitting
(10\% training, 10\% validation, and 80\% testing) in both homophilic and heterophilic datasets. We tune the hyperparameters such as learning rate and weight decay for GPRGNN following the original paper~\cite{chien2021GPR-GNN}. We run the experiments for 10 random data splits and report the mean and standard deviation.
The comparison in Figure~\ref{fig:main} shows the following observations:
\vspace{-0.1in}
\begin{itemize}[leftmargin=0.2in]
\item On homophilic datasets, i.e., Figure~\ref{fig:main}(a), GPRGNN underperforms GPRGNN-optimal on Cora and CiteSeer, and they achieve comparable performance on PubMed and Computers. Both of them outperform MLP significantly on all datasets. This demonstrates that GPRGNN can learn effective low-pass filters to model the graph signal for homophilic graphs, but it does not fully reach its maximum potential and is notably inferior to the optimal polynomial filter (GPRGNN-optimal) in some cases.

\item On heterophilic datasets, i.e., Figure~\ref{fig:main}(b),  GPRGNN exhibits significantly worse performance than
GPRGNN-optimal, indicating that GPRGNN learns suboptimal filters. 
Moreover, GPRGNN does not show notably better performance than MLP, and it even underperforms MLP significantly on Cornell and Texas datasets while GPRGNN-optimal is consistently superior to MLP.
This demonstrates that GPRGNN struggles to model the complex graph signal in heterophilic graphs and is far from achieving its theoretical learning potential as indicated by GPRGNN-optimal. 

\item GPRGNN exhibits unstable performance and high randomness. The learning process of GPRGNN is heavily influenced by the randomness of weights initialization, data splitting and optimization process, leading to a large variance of the final results. 

\end{itemize}

These observations reveal that GNNs based on the polynomial filters indeed have great potential, but the current learning approach is suboptimal, which prevents them from showcasing their optimal capacity. However, the reasons for such failures are unclear.

\subsection{Overfitting issue} 
In this section, we provide further insight into the failures of polynomial filter learning in GNNs and reveal one of the key obstacles to their success.  
Specifically, we compare the training process of 
GPRGNN and GPRGNN-optimal by measuring their training, validation, and testing accuracy during the training process. We adopt the same settings in Section~\ref{sec:search-optimal}, and only run one split. The training process as shown in Figure~\ref{fig:figure2} provides the following observation:
\begin{itemize}[leftmargin=0.2in]
\item GPRGNN always achieves perfect training accuracy close to $100\%$ on all datasets, but its validation performance and test performance are significantly lower. This reveals that the polynomial filter is flexible enough so that the training data can be perfectly fitted but the big generalization gap does not deliver satisfying performance on validation and test data.

\item GPRGNN-optimal achieves lower training accuracy than GPRGNN, but its validation and test performance are notably better than GPRGNN. This indicates that GPRGNN suffers from a severe overfitting issue while the optimal polynomial filter can mitigate the overfitting and narrow down this generalization gap.
\end{itemize}

To summarize, polynomial filter learning in GNNs suffers from a severe overfitting problem that leads to poor generalization. This provides a plausible explanation for the failures of polynomial-based GNNs regardless of their powerful expressiveness.

\section{Auto-Polynomial}
\label{sec:alg}

In this section, we introduce a novel and general automated polynomial filter learning approach to address the aforementioned overfitting issues and improve the effectiveness of graph polynomial filters.

\subsection{Automated polynomial filter learning}

The polynomial coefficients in polynomial filters serve as key parameters that have a strong influence on the prediction of GNNs. Existing polynomial-based GNNs such as ChebNet, GPRGNN, and BernNet treat the polynomial coefficients the same as other model parameters, both of which are jointly learned through gradient descent algorithms such as Adam~\cite{kingma2014adam} according to the training loss. Our preliminary study in Section~\ref{sec:overfit} reveals the suboptimal of the widely used polynomial filter learning approach. This suboptimality is partially due to the severe overfitting problem since the filters are adjusted to perfectly fit the labels of training nodes that have a low ratio in the typical semi-supervised learning settings for graph deep learning~\cite{yang2016revisiting,kipf2016semi}. On the other hand, our brute-force search on the polynomial filter based on the validation accuracy achieves consistently better performance and mitigates the overfitting issues. However, when the polynomial order $K$ is large, it is not plausible to brute-force search the optimal polynomials by trial and error. 
This motivates us to develop a better automated learning approach for polynomial graph filter learning that mitigates the overfitting.

\vspace{0.05in}
\noindent \textbf{Auto-Polynomial.} 
We propose Auto-Polynomial, a novel automated polynomial filter learning approach that learns the polynomial coefficients guided by the validation loss inspired by hyperparameter optimization (HPO) and automated machine learning (AutoML).
Instead of treating polynomial coefficients the same as other model parameters, we consider these polynomial coefficients as hyperparameters.
Let the polynomial coefficients $\Theta=\{\theta_k\}_{k=0}^K$ represent GNNs with different filtering properties and $w$ denotes other learnable weights in GNNs.
The main idea behind Auto-Polynomial is
to jointly optimize the coefficients $\Theta$ and the weights $w$ but with totally different targets.
Formally, we model the automated polynomial filter learning as a bi-level optimization problem:
 \begin{equation}
 \label{eq:bi-opt}
    \begin{aligned}
    \underset{\Theta  }{\mathrm {min} }&\quad \mathcal{L} _{val} (w^{*} (\Theta ),\Theta )\\
    \mathrm {s.t.}&\quad w^{*} (\Theta )   =\underset{w}{\mathrm {argmin}} (\mathcal{L}_{train}(w,\Theta ) )
    \end{aligned}
 \end{equation}
where the lower-level optimization problem finds the best model parameters $w^*(\Theta)$ based on training loss while the upper-level problem finds the optimal polynomial filter represented by $\Theta$ based on validation loss.
Guided by this bi-level optimization problem, the learned GNN model can not only fit the training data accurately but also enhance its performance on validation data, which helps improve generalization ability and alleviate the overfitting problem.

\vspace{0.05in}
\noindent \textbf{Approximate Computation.}
The bi-level problem can be solved by alternating the optimization between lower-level and upper-level problems. However, the optimal solution $w^*(\Theta)$ in the lower-level problem has a complicated dependency on polynomial coefficients $\Theta$ so that the meta-gradient of upper-level problem, i.e., 
$\bigtriangledown_{\Theta }\mathcal{L} _{val} (w^{*} (\Theta ),\Theta )$,
is hard to compute. This complexity becomes even worse when the lower-level problem adopts multiple-step iterations.
Therefore, we adopt a simplified and elegant method inspired by DARTS~\cite{liu2018darts} to approximate the gradient computation in problem~\eqref{eq:bi-opt}:
\begin{equation}
\label{eq:darts}
    \bigtriangledown_{\Theta }\mathcal{L} _{val} (w^{*} (\Theta ),\Theta )
\approx \bigtriangledown_{\Theta }\mathcal{L}_{val}(w-\xi \bigtriangledown_{w}\mathcal{L}_{train}(w,\Theta),\Theta)
\end{equation}
where $\xi$ denotes the learning rate for the lower-level optimization, and only one step gradient is used to approximate the optimal model $w^{*}(\Theta)$. This approximation provides a reasonable solution since $w$ will accumulate the update from all previous training iterations and its initial value $w$ might not be too far away from the optima.  

In particular, if we set $\xi=0$, Equation~\eqref{eq:darts} is equivalent to updating $\Theta$ based on the current weights $w$, which is a first-order approximation, a much simpler form of the problem. Although first-order approximation may not provide the best result, it can improve computational efficiency. If we use $\xi>0$, then we can use chain rule to approximate Equation~\eqref{eq:darts}:
\begin{equation}
    \label{eq:chain}
    \nabla_{\Theta} \mathcal{L}_{v a l}\left(w^{\prime}, \Theta\right)-\xi \nabla_{\Theta, w}^{2} \mathcal{L}_{t r a i n}(w, \Theta) \nabla_{w^{\prime}} \mathcal{L}_{v a l}\left(w^{\prime}, \Theta\right)
\end{equation}
where $w^{\prime}=w-\xi \nabla_{w} \mathcal{L}_{\text {train }}(w, \Theta)$ denotes the weights after one step gradient descent. In practice, we can apply the finite difference method~\cite{liu2018darts} to approximate the 
second term in Equation~\eqref{eq:chain}:
\begin{equation}
\label{eq:2-order}
\begin{aligned}
    &\nabla_{\Theta, w}^{2} \mathcal{L}_{\text {train }}(w, \Theta) \nabla_{w^{\prime}} \mathcal{L}_{val }\left(w^{\prime}, \Theta\right) \\
\approx &\frac{\nabla_{\Theta} \mathcal{L}_{train }\left(w^{+}, \Theta\right)-\nabla_{\Theta} \mathcal{L}_{train }\left(w^{-}, \Theta\right)}{2 \varepsilon}
\end{aligned}
\end{equation}
where $\varepsilon$ denotes a small scalar in finite difference approximation, $w^{+}=w + \varepsilon \nabla_{w^{\prime}} \mathcal{L}_{v a l}\left(w^{\prime}, \Theta\right)$, and $w^{-}=w - \varepsilon \nabla_{w^{\prime}} \mathcal{L}_{v a l}\left(w^{\prime}, \Theta\right)$. 
In practice, we also introduce a hyper-parameter $freq$ that specifies the update frequency of polynomial coefficients to further improve the efficiency by skipping this computation from time to time. 

\subsection{Use cases}
The proposed Auto-Polynomial is a general polynomial filter learning framework and it can be applied to various polynomial filter-based GNNs. In this section, we will introduce some representative examples to showcase its application.

\vspace{0.05in}
\noindent \textbf{Case 1: Auto-GPRGNN.}
In GPRGNN~\cite{chien2021GPR-GNN}, the weights of Generalized PageRank (GPR), i.e., polynomial coefficients $\Theta$ in Equation~\eqref{eq:GPRGNN}, are trained together with the model parameters $w$. The GPR weights $\Theta$ can be adaptively learned during the training process to control the contribution of the node features and their propagated features in different aggregation layers.
GPR provides a flexible way to model various graph signals.
However, our preliminary study in Section~\ref{sec:overfit} has shown that GPRGNN can not learn effective filters and faces a serious overfitting problem for the learning of polynomial coefficients. Moreover, its performance is very sensitive to the initialization of filters, data split, and random seeds.
The proposed Auto-Polynomial framework can help mitigate these problems. 
Instead of learning two types of parameters simultaneously, we treat the training of the model parameter $w$ and the GPR weight $\Theta$ as the lower-level and upper-level optimization problems in Equation~\eqref{eq:bi-opt}, respectively.
In this work, we denote GPRGNN using Auto-Polynomial as Auto-GPRGNN.

\vspace{0.05in}
\noindent \textbf{Case 2: Auto-BernNet.}
In BernNet~\cite{he2021bernnet}, the coefficients of the Bernstein basis $\Theta$ can be learned end-to-end to 
fit any graph filter in the spectral domain. The training process of BernNet is slightly different from GPRGNN 
since BernNet specifies an independent learning rate for the polynomial coefficients. To some extent, the learning of polynomial coefficients $\Theta$ is slightly distinguished from the learning of model parameters $w$, which improves the model performance.
However, this learning method still has not really brought out the potential of BernNet, and faces the problem of overfitting, especially under semi-supervised learning on heterophilic data.
Applying the proposed Auto-Polynomial method to BernNet can ease this problem. 
We can regard the Bernstein basis coefficients $\Theta$ in BernNet as hyperparameters and optimize them in the Auto-Polynomial bi-level optimization problem. Analogously, we refer to BernNet with Auto-Polynomial as Auto-BernNet.

\begin{algorithm}[htb]

  \SetAlgoLined
  \KwIn{
    \textbf{$\Theta$}: coefficients of polynomial filter \\
    \textbf{$\xi$}: learning rate for the lower-level approximation \\
    $\eta_{0}$: learning rate of polynomial filter \\
    $\eta_{1}$: learning rate of model parameters \\
    \textbf{$w$}: model parameters,\\
    \textbf{$freq$}: frequency to update $\Theta$
  }
  \KwOut{Graph polynomial filter with coefficients $\Theta$}
  \For{$i \leftarrow 1$ \KwTo 
  $Iteration$
  }{
    (\textbf{Update $\Theta$}):
    
   \eIf{i \% freq = 0}{
       
    $w^{\prime}_{i} \leftarrow w_{i} - \xi \nabla_{w_{i}} \mathcal{L}_{train}(w_{i}, \Theta_{i})$,

    $w^{+}_{i}\leftarrow w_{i}+\varepsilon \nabla_{w^{\prime}_{i}} \mathcal{L}_{v a l}\left(w^{\prime}_{i}, \Theta_{i}\right)$,
    
    $w^{-}_{i}\leftarrow w_{i}-\varepsilon \nabla_{w^{\prime}_{i}} \mathcal{L}_{v a l}\left(w^{\prime}_{i}, \Theta_{i}\right)$,

    $\nabla_{\Theta_i, w_i}^{2} \mathcal{L}_{ train }(w_i, \Theta_i) \nabla_{w_i^{\prime}} \mathcal{L}_{val }\left(w_i^{\prime}, \Theta_i\right) \leftarrow \text{Equ.}~\eqref{eq:2-order}$

    $\nabla_{\Theta_{i}} \mathcal{L}_{v a l}\left(w^{*}_{i}(\Theta_{i}), \Theta_{i}\right) \leftarrow \text{Equ.}~\eqref{eq:chain}$
    
    $\Theta_{i+1} \leftarrow \Theta_{i} - \eta_{0} \nabla_{\Theta_{i}} \mathcal{L}_{val}(w^{*}_{i}, \Theta_{i})$}{$\Theta_{i+1} \leftarrow \Theta_{i}$
    }
   
     (\textbf{Update $w$}):
     $w_{i+1}=w_{i}-\eta_{1} \nabla_{w_{i}} \mathcal{L}_{train}(w_{i}, \Theta_{i+1})$;
  }
  \caption{Auto-Polynomial Learning Algorithm}
  \label{alg:auto-poly}
\end{algorithm}

\subsection{Implementation}
In this section, we demonstrate the implementation details of Auto-Polynomial and present the learning procedure in Algorithm~\ref{alg:auto-poly}.

\vspace{0.05in}
\noindent \textbf{Bi-level update.}
The whole procedure of Auto-Polynomial includes two main steps: updating the  polynomial coefficients $\Theta$ and updating the model parameters $w$.
To update $\Theta$,  we perform gradient descent with
$\nabla_{\Theta} \mathcal{L}_{val}$ to enhance the generalization ability of polynomial filter learning.
To reduce memory consumption and increase efficiency, we first employ the finite difference
method to compute the second-order term in Equation~\eqref{eq:2-order} and then obtain an approximation of $\nabla_{\Theta} \mathcal{L}_{val}$ as Equation~\eqref{eq:chain}.
In practice, we introduce a hyper-parameter $freq$ to specify the update frequency of polynomial coefficients, so as to further improve the algorithm efficiency in necessary. In other words, $\Theta$ can be updated only once in every $freq$ iterations of model update.
To update $w$,  we utilize
$\nabla_{\Theta} \mathcal{L}_{train}$ to update the model parameters $w$ by fitting the labels of training nodes. Through this alternating update process, we ultimately obtain appropriate polynomial coefficients and model parameters to improve the GNNs' graph modeling and generalization abilities.

\vspace{0.05in}
\noindent \textbf{Complexity analysis.}
The bi-level optimization problem in Auto-Polynomial can be solved efficiently. 
To be specific, we assume that the computation complexity of the original model is $O(C)$, where $C$ is a known constant. Then the model complexity after applying Auto-Polynomial is $O(\frac{freq+4}{freq}C) $, as we need $4$ extra forward computation when updating $\Theta$. Therefore, the computational complexity of Auto-Polynomial is at most 5 times bigger than the backbone GNN model but it can be effectively reduced by choosing a larger $freq$. In terms of memory cost, Auto-Polynomial does not increase the memory cost significantly no matter how large $freq$ is. The computation time and memory cost will be evaluated in Section~\ref{sec:exp}. 

\section{Experiment}
\label{sec:exp}

In this section, we provide comprehensive experiments to validate the effectiveness of Auto-Polynomial under semi-supervised learning and supervised learning settings on both heterophilic and homophilic datasets. Further ablation studies are presented to investigate the influence of  training set ratio and update frequency on the model performance.

\subsection{Baselines and Datasets}
\label{sec:exp-set}
\vspace{0.05in}
\noindent \textbf{Baselines.}
We compare the proposed Auto-Polynomial, including use cases Auto-GPRGNN and Auto-BernNet, with 7 baseline models: MLP, GCN~\cite{kipf2016semi}, ChebNet~\cite{Chebnet}, APPNP~\cite{appnp}, GPRGNN~\cite{chien2021GPR-GNN}, BernNet~\cite{he2021bernnet} and H2GCN~\cite{zhu2020beyond}. 
For GPRGNN\footnote{https://github.com/jianhao2016/GPRGNN} and BernNet\footnote{https://github.com/ivam-he/BernNet}, we use their the officially released code. For H2GCN, we use the Pytorch version of the original code. 
For other models, we used the implementation of the Pytorch Geometric Library~\cite{fey2019fast}.

\vspace{0.05in}
\noindent \textbf{Datasets.}
 We conduct experiments on the most commonly used real-world benchmark datasets for node classification. We use 4 homophilic benchmark datasets, including three citation graphs including Cora, Citeseer and PubMed~\cite{sen2008collective,namata2012query}, and the Amazon co-purchase graph Computers~\cite{mcauley2015image}.
 We also use 4 heterophilic benchmark datasets, including Wikipedia graphs such as Chameleon and Squirrel\cite{rozemberczki2021multi}, and webpage graphs such as Texas and Cornell from WebKB\footnote{http://www.cs.cmu.edu/afs/cs.cmu.edu/project/theo-11/www/wwkb}.
 We summarize the dataset statistics in Table~\ref{tab:stat-data}.

\vspace{0.05in}
\noindent \textbf{Hyperparameter settings.}
For APPNP, we set its propagation step size $K=10$ and optimize the teleport probability $\alpha$ over  $\left \{ 0.1, 0.2, 0.5, 0.9\right \}.$
 For ChebNet, we set the propagation step $K=2$.
 For GPRGNN, we set its polynomial order $K=10$.
For BernNet, we set the polynomial order $K=10$ and tune the independent learning rate for propagation layer  over $\left \{ 0.002, 0.005, 0.01, 0.05\right \}.$
For H2GCN, we follow the settings in~\cite{zhu2020generalizing}. 
We search the embedding round $K$ over $\left \{ 1, 2\right \}.$
For all models except H2GCN, we use 2 layer neural networks with
64 hidden units and set the dropout rate 0.5.
For Auto-GPRGNN and Auto-BernNet, we use 2-layer MLP with 64 hidden units and set the polynomial order $K=10$. 
We search the meta learning rate $\eta_0$ of Auto-Polynomial over $\left \{ 0.01, 0.05\right \}$ and weight decay over $\left \{0, 0.0005\right \}$. We search the learning rate $\xi$ over $\left \{0, 0.05\right \}$ and the polynomial weights are randomly initialized.
For all models, we adopted an early stopping strategy of 200 epochs with a maximum of 1000 epochs. We use the Adam optimizer to train the models. We optimize learning rate over $\left\{0.002, 0.01, 0.05\right\}$, and weight decay over $\left\{0, 0.0005\right\}$.
We select the best performance according to the validation accuracy.

 \begin{table*}[htbp]
    \centering
    \caption{Statistics of real-world datasets}
    \vspace{-0.1in}
    \label{tab:stat-data}
    \begin{tabular}{ccccccccc}
        \hline
        Statistics &Cora & Citeseer &Pubmed& Computers &Chameleon& Cornell & Squirrel&Texas \\
        \hline
        Features & 1433& 3703& 500& 767& 2325&  1703& 2089& 1703\\
         Nodes & 2708& 3327& 19717& 13752& 2277&  183& 5201& 183 \\
        Edges& 5278& 4552& 44324& 245861& 31371&  277&198353& 279\\
        Classes& 7& 6& 5& 10& 5& 5& 5& 5\\
        $h$ & 0.83& 0.72& 0.8& 0.8& 0.25& 0.3& 0.22& 0.06\\
        \hline
    \end{tabular}
    \vspace{0.1in}
\end{table*}

\subsection{Semi-supervised node classification}

\begin{table*}[!htp]
    \centering
    \caption{Semi-supervised learning: mean accuracy ± stdev. The best (second best) results are marked in boldface (underlined).}
    \vspace{-0.1in}
    \resizebox{\linewidth}{!}{%
    \begin{tabular}{cccccccccc}
        \hline
        &MLP & ChebNet &GCN& APPNP &H2GCN& GPRGNN & BernNet&Auto-GPRGNN&Auto-BernNet \\
        \hline
        Cora &64.79±0.70&	81.37±0.93&	83.11±1.05&	\textbf{84.71±0.53}&82.84±0.69	&	84.37±0.89&	84.27±0.81&	\underline{84.71±1.19}&	84.66±0.60\\
        Citeseer &64.90±0.94&	71.56±0.86&	71.67±1.31&	72.70±0.80&	72.61±0.67&72.17±0.71&	72.08±0.63&	\underline{72.76±1.18}&\textbf{72.85±0.94}	\\
        PubMed& 84.64±0.41&	87.17±0.31	&86.40±0.22&	86.63±0.36&	86.71±0.25&	86.42±0.34&	\textbf{86.94±0.18}	&86.69±0.31&\underline{86.81±0.41}\\
        Computers& 76.80±0.88&	86.17±0.58&	86.79±0.63&	85.76±0.44&84.99±0.54	&	85.79±1.00	&\underline{86.84±0.57}	&86.74±0.61	&\textbf{87.61±0.48}\\
        Chameleon &38.61±1.04&	49.14±1.66&	51.65±1.35&	42.13±2.06&50.69±1.60	&41.25±4.44	&54.11±1.16	&\underline{55.34±2.93}&	\textbf{56.12±1.13}	\\
        Cornell &\underline{66.28±5.97}&	54.21±5.90&	48.83±5.42&	63.38±5.28&63.52±6.38	&42.83±5.92&	63.79±5.30&	58.55±7.44	&\textbf{67.03±4.49}	\\
        Squirrel& 26.58±1.37&	32.73±0.82&	\textbf{36.98±1.03}&	29.70±1.31&29.69±0.80	&	28.83±0.50	&26.75±4.49&	33.16±2.83&	\underline{34.48±1.29}\\
        Texas &71.62±4.23&	70.74±5.10&	58.18±5.11&	68.45±7.35&\underline{75.00±3.42}	&	41.96±12.59&	71.62±5.26	&73.85±2.64	&\textbf{75.47±3.18}\\
        \hline
    \end{tabular}
    } 
    \label{tab:semi}
\end{table*}

\noindent \textbf{Experimental settings.}
In semi-supervised learning setting, we randomly split the datasets into training/validation/test sets with the ratio of  10\%/10\%/80\%. We run each experiment 10 times with random splits and report the mean and variance of the accuracy. 
Note that the data split for semi-supervised learning in the literature is quite inconsistent since existing works usually use different ratios for the homophilic and heterophilic datasets~\cite{chien2020adaptive}, or a higher fraction on both the homophilic and heterophilic datasets~\cite{pei2020geom,chen2020simple, he2021bernnet}. 
To be fair, we use a consistent ratio for both the homophilic and the heterophilic datasets, which can truly reflect the adaptability of the polynomial filters on different datasets.

\vspace{0.05in}
\noindent \textbf{Performance analysis.}
The performance summarized in Table~\ref{tab:semi} shows the following major observations: 
\begin{itemize}[leftmargin=0.2in]
\item 
For homophilic datasets, all GNN models outperform MLP significantly, indicating that the structure information of homophilic graphs can be learned and captured easily. Moreover, Auto-GPRGNN and Auto-BernNet can achieve the best performance in most cases, demonstrating Auto-Polynomial's superior capability in polynomial graph filter learning.  
\item For heterophilic datasets, some baselines such as GCN, ChebNet and GPRGNN fail to learn effective polynomial filters and they even underperform graph-agnostic MLP. 
However, Auto-GPRGNN and Auto-BernNet can outperform all baselines by a significant margin. For instance, Auto-GPRGNN improves over GPRGNN by 14\%, 16\%, 5\%, and 31\% on Chameleon, Cornell, Squirrel, and Texas datasets, respectively.
Auto-BernNet improves over BernNet by 2\%, 4\%, 8\%, and 4\% on Chameleon, Cornell, Squirrel, and Texas datasets, respectively.
Moreover, Auto-polynomial effectively reduces the standard deviations in most cases. Especially, Auto-GPRGNN archives a standard deviation 10\% lower than GPRGNN on the Texas dataset.
These comparisons clearly indicate that Auto-Polynomial is 
capable of learning polynomial filters more effectively and mitigating the overfitting issue.
\end{itemize}

\subsection{Supervised node classification}

\begin{table*}[!htp]
\vspace{0.1in}
    \centering
    \caption{Supervised learning: mean accuracy ± stdev. 
    The best (second best) results are marked in boldface (underlined).}
    \vspace{-0.1in}
    \resizebox{\linewidth}{!}{%
    \begin{tabular}{cccccccccc}
        \hline
        &MLP & ChebNet &GCN& APPNP &H2GCN& GPRGNN & BernNet&Auto-GPRGNN&Auto-BernNet \\
        \hline
        Cora& 76.20±1.83&	86.67±1.99&	86.59±1.31&	88.70±0.93	&87.36± 1.11&	\underline{88.79±1.37}&	86.53±1.26&	\textbf{89.05±1.09}&	88.70±0.93 \\
        Citeseer& 74.76±1.26&	77.44±1.77	&77.43±1.42	&79.43±1.75	&\textbf{82.78±1.21}&	78.95±1.93&	78.58±1.82&	79.76±0.89&	\underline{79.82±1.55}\\
        PubMed& 86.51±0.82&	88.76±0.58&	87.04±0.60&	\underline{89.10±0.89}&	87.72±0.46&	88.09±0.51&	88.25±0.52&\textbf{89.10±0.54}	&	88.38±0.45\\
        Computers& 83.07±0.74&	\textbf{89.11±0.61}&	87.71±0.73&	86.58±0.44&\underline{89.04±0.61}	&	87.04±2.71&	86.55±0.54&	88.56±0.54&	88.71±0.71\\
        Chameleon&47.62±1.41&	60.68±2.22&	61.90±2.19&	52.65±2.48&59.58±2.07	&	62.54±2.77&	65.76±1.57&\underline{67.38±1.44}&	\textbf{68.07±1.48}\\
        Cornell& 87.03±3.78&	77.57±6.05&	60.27±11.53&	88.65±5.64&	87.36±5.82&	87.03±4.95&	83.24±4.32&	\textbf{91.08±5.28}	&\underline{89.73±3.15}\\
        Squirrel &31.16±1.41&	40.46±1.81&	44.38±1.78&	35.86±1.54&35.78±1.31	&	46.55±1.72&	\textbf{49.30±1.81}&	47.02±2.54&	\underline{47.41±1.41}\\
        Texas& 89.02±2.5&	85.88±4.45	&73.33±6.09&	87.65±1.97&88.24±4.64	&	81.76±15.8&	88.63±3.37&	\underline{89.41±3.42}&	\textbf{90.39±2.83}\\
        \hline
    \end{tabular}
    }
    \label{tab:full}
\end{table*}

\begin{figure*}[!htp] 
\vspace{0.1in}
  \centering \subcaptionbox{Chameleon\label{fig:subfig2}}{\includegraphics[width=0.26\linewidth]{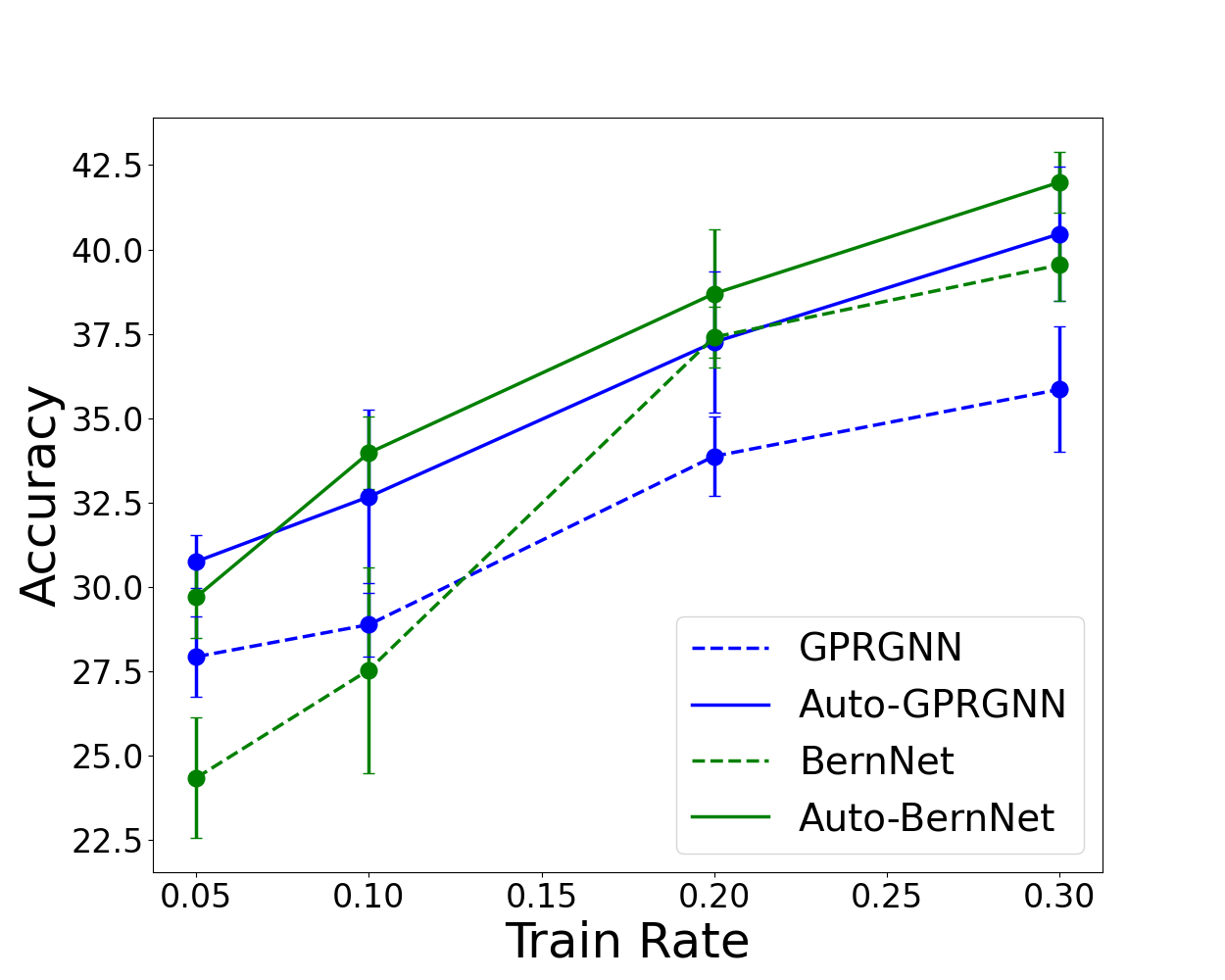}}
  \hspace{-15pt}
  \subcaptionbox{Squirrel\label{fig:subfig3}}{\includegraphics[width=0.26\linewidth]{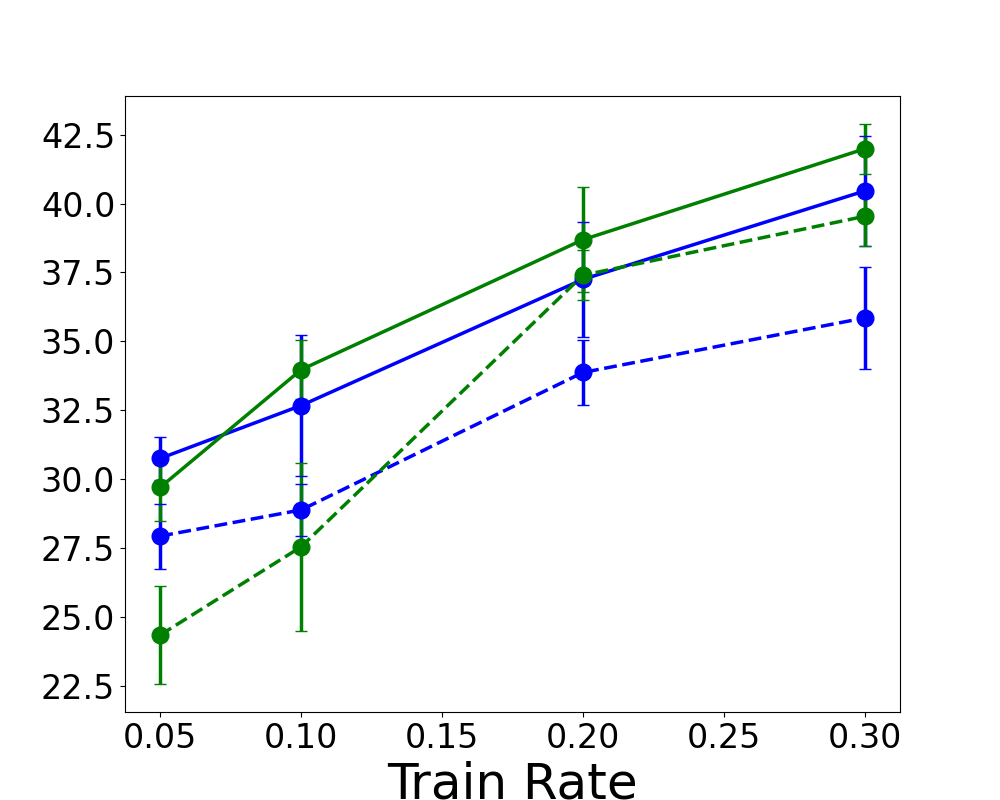}}
  \hspace{-15pt}
  \subcaptionbox{Texas\label{fig:subfig4}}{\includegraphics[width=0.26\linewidth]{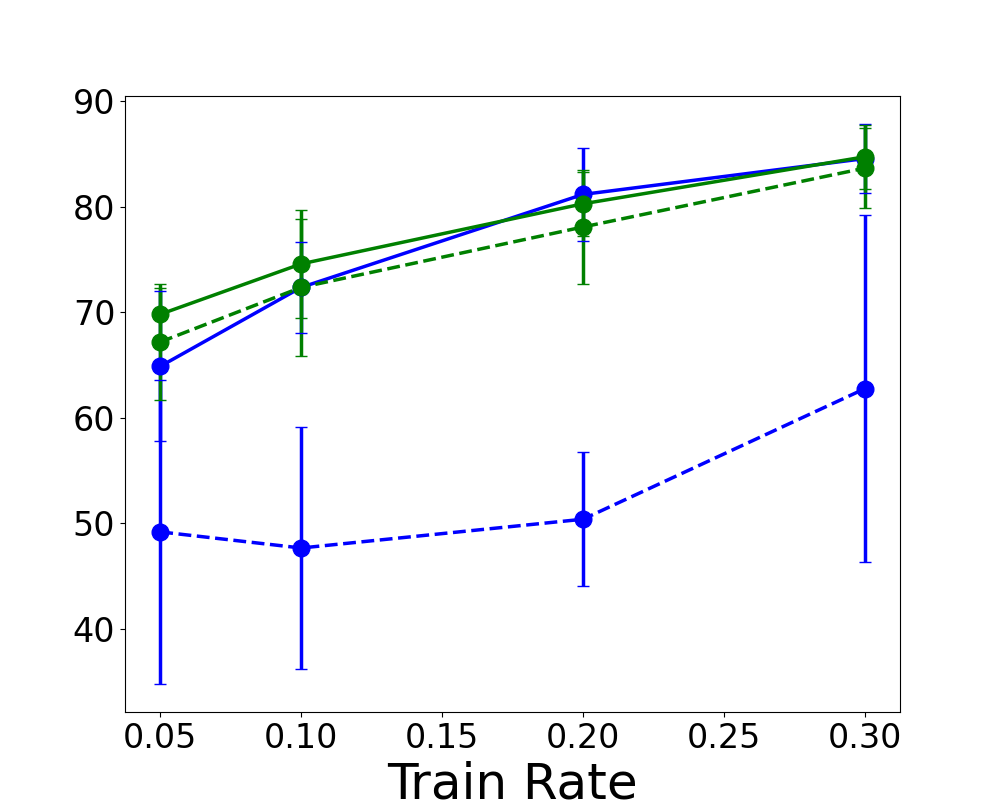}}
  \hspace{-15pt}
  \subcaptionbox{Cora\label{fig:subfig1}}{\includegraphics[width=0.26\linewidth]{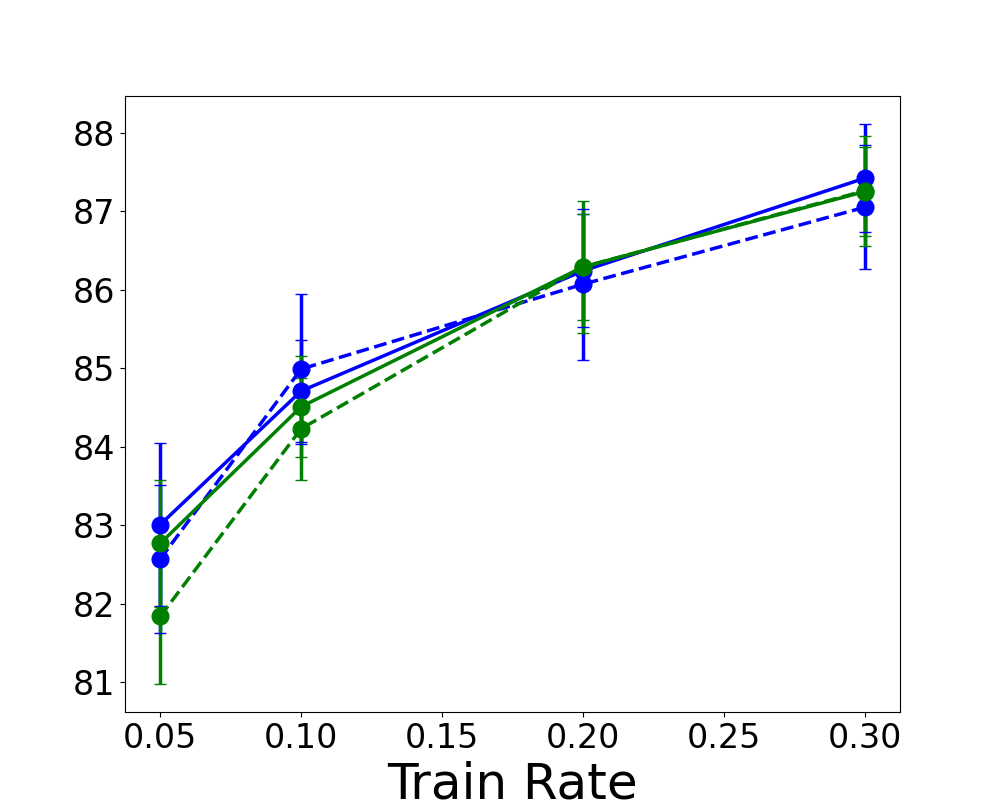}}
  \hspace{-15pt}
  \vspace{-0.1in}
  \caption{Results under different ratios on 3 heterophilic datasets and 1 homophilic dataset.  
  }
  \label{fig:figure3}
\end{figure*}

\vspace{0.05in}
\noindent \textbf{Experimental settings.}
In supervised learning setting, 
we  randomly split the datasets into training/validation/test sets with the ratio of  48\%/32\%/32\%, following the high ratio in the work~\cite{zhu2020beyond}. We run each experiment 10 times with random
splits and report the mean and variance of the test accuracy.

\vspace{0.05in}
\noindent \textbf{Performance analysis.} The performance summarized in Table~\ref{tab:full} shows the following major observations:
\begin{itemize}[leftmargin=0.2in]
\item 
For homophilic datasets, most of the models performed well with small standard deviations, which shows that the existing GNNs are capable of learning proper filters for homophilic graphs. Moreover, Auto-GPRGNN and Auto-BernNet improves their backbone models, GPRGNN and BernNet, indicating that Auto-Polynomial helps them learn better polynomial filters.

\item For heterophilic datasets, most of the baselines perform poorly with large standard deviations, which shows that they cannot learn heterophilic graph information well. Models using polynomial filters such as ChebNet, GPRGNN, and BernNet achieve better performance, while Auto-GPRGNN and Auto-BernNet achieve the best or second-best performance in most cases and their standard deviations are smaller in most cases. Especially, Auto-GPRGNN achieves a standard deviation 12\% lower than GPRGNN on the Texas dataset.
These results indicate the effectiveness of Auto-Polynomial in graph polynomial filter learning.

\end{itemize}

\begin{table*}[!htp]
\vspace{0.2in}
\centering
\caption{Efficiency  
of GPRGNN and Auto-GPRGNN on Cora and Chameleon. $freq$ denotes the filter update frequency. 
}
\vspace{-0.1in}
\label{tab:ablation-freq}
\resizebox{\linewidth}{!}{%
 \begin{tabular}{ccccccc}
 \hline
  &   \multicolumn{3}{c}{Cora} & \multicolumn{3}{c}{Chameleon}  \\
\cmidrule(r){2-4}  \cmidrule(r){5-7}
 \noalign{\smallskip} 
 Method & Test Acc & Running Time (S) & Memory Cost (MB) & Test Acc & Running Time (S) & Memory Cost (MB) \\ 
   \hline
 GPRGNN &84.70±0.47&	1.04&49.67 &40.03±4.31	&1.13&64.43 \\
 Auto-GPRGNN ($freq$=1)  & 84.35±0.47 &4.18&66.94 &54.96±2.29&	5.92&88.29\\ 
 Auto-GPRGNN ($freq$=2)    & 84.51±1.03 &2.71&66.94 &53.71±1.89&	3.79&88.29\\ 
 Auto-GPRGNN ($freq$=3) & 84.65±0.73&2.25& 66.94&52.71±3.46	&3.38&88.29\\
 Auto-GPRGNN ($freq$=4) & 84.76±0.96&2.00& 66.94&53.23±3.37&	2.90&88.29\\
 Auto-GPRGNN ($freq$=5) &83.94±0.79&1.91& 66.94&52.91±2.35	&2.66&88.29\\
   \hline
 \end{tabular}
 } 
\end{table*}

\subsection{Ablabtion study}

In this section, we provide ablation studies on the labeling ratio and polynomial update frequency.

\vspace{0.05in}
\noindent \textbf{Labeling ratio.}
From the results in semi-supervised and supervised settings in Table~\ref{tab:semi} and Table~\ref{tab:full}, it appears that the improvement from Auto-Polynomial is more significant on semi-supervised tasks (low labeling ratio), as compared to supervised tasks (high labeling ratio). This phenomenon motivates the study of how the labeling ratio impacts the effectiveness of our proposed framework. 
To this end, we fix the validation set ratio at 10\%, vary the training set ratio over $\left\{5\%,\ 10\%,\ 20\%,\ 30\%\right\}$, and adjust the  test set ratio accordingly.
We evaluate the performance on three representative heterophilic datasets (Chameleon, Squirrel and Texas), and one homophilic dataset (Cora). We compare the proposed Auto-GPRGNN and Auto-BernNet with the original GPRGNN and BernNet, following the hyperparameter settings in Section~\ref{sec:exp-set}. The results under different ratios on four datasets are shown in Figure~\ref{fig:figure3}, and we can have the following observations:
\begin{itemize}[leftmargin=0.2in]
    \item Auto-GPRGNN and Auto-BernNet 
    can achieve significant improvements in heterophilic graphs compared to the original GPRGNN and BernNet. This suggests that the Auto-Polynomial framework can help the polynomial filters effectively adapt to heterophilic graphs. 
    \item The improvements of Auto-Polymonial over the backbone models are even more obvious under lower labeling ratios. This shows that when there are not enough labeled samples available, the proposed Auto-Polynomial framework can offer greater advantages in addressing the overfitting issues and enhancing the model's generalization ability.

\end{itemize}

\vspace{0.05in}
\noindent \textbf{Update frequency.}
In practical implementation, we introduce a hyper-parameter $freq$ to control the update frequency of polynomial coefficients and thus improve the efficiency of our models. To investigate the influence of update frequency on the efficiency and performance of Auto-Polynomial, we vary the frequency at which Auto-GPRGNN updates $\Theta$ and examine the the resulting changes in performance, training time and memory cost.
We conduct the experiments on both homophilic dataset (Cora) and  heterophilic dataset (Chameleon). The results presented in Table~\ref{tab:ablation-freq} provide the following observations: 
\begin{itemize}[leftmargin=0.2in]
    \item On the homophilic dataset (Cora), reducing the update frequency of Auto-GPRGNN almost does not impact the performance, 
    validating the stability of Auto-Polynomial.
    \item On the heterohpilic dataset (Chameleon), Auto-GPRGNN  outperforms GPRGNN by  significantly and is not very sensitive to the update frequency. This indicates that
    Auto-Polynomial can greatly enhance the model's efficiency by using an appropriate update frequency, with negligible sacrifice in performance.
    \item The memory usage of Auto-GPRGNN is approximately 1.3 times that of GPRGNN. However, this increase in memory usage can be deemed acceptable, given the significant improvement achieved by Auto-GPRGNN.
    
\end{itemize}

To summarize, Auto-Polynomial provides an effective, stable, and efficient framework to improve the performance and generalization of polynomial filter learning.

\section{Related Work}

\vspace{0.05in}
\noindent \textbf{Polynomial Graph Filter.}
Polynomial graph filters have been widely used as the guiding principles in the design of spectral-based GNNs starting from the early development of ChebNet~\cite{Chebnet}. It has gained increasing attention recently due to its powerful flexibility and expressiveness in modeling graph signals with complex properties. For instance, some representative works including ChebNet~\cite{Chebnet},
GPRGNN~\cite{chien2021GPR-GNN} and BernNet~\cite{he2021bernnet} utilize Chebyshev basis, Monomial basis,
and Bernstein basis, respectively, to 
approximate the polynomial filters whose coefficients can be learned adaptively to model the graph signal. Therefore, these polynomial-based approaches exhibit encouraging results in graph signal modeling on both homophilic and heterophilic graphs.
In this work, we design novel experiments to explore the potential and limitations of graph polynomial learning approaches, and we propose Auto-Polynomial, a general automated learning framework to improve the effectiveness of any polynomial-based GNNs, making this work a complementary effort to existing works.

\vspace{0.05in}
\noindent \textbf{AutoML on GNNs.}
Automated machine learning (AutoML)~\cite{hutter2019automated} has gained great attention during the past few years due to its great potential in automating various procedures in machine learning, such as data augmentation~\cite{cubuk2018autoaugment, Cubuk_2020_CVPR_Workshops}, neural architecture searching (NAS)~\cite{zoph2016neural,liu2018darts}, and hyper-parameter optimization (HPO)~\cite{franceschi2018bilevel,feurer2019hyperparameter,pmlr-v108-lorraine20a}.
Recent works have applied AutoML to GNN architecture search using various search strategies such as random search~\cite{you2020design,gao2019graphnas,gao2021graph,tu2019autone},
reinforcement learning~\cite{gao2021graph,zhou2019auto,lu2020fgnas}, evolutionary algorithms~\cite{nunes2020neural,shi2020evolutionary,guan2021autoattend}, and 
differentiable search~\cite{zhao2020probabilistic,zhao2021efficient,zhao2020learned}. 
However, these works mainly focus on the general architecture search and none of them provides an in-depth investigation into the polynomial filter learning. Moreover, these methods require a vast search space and complex learning algorithms, which can be time-consuming and resource-intensive. Different from existing works, this work provides a dedicated investigation into polynomial graph filter learning and proposes the first automated learning strategy to significantly and consistently improve the effectiveness of polynomial filter learning with a highly efficient learning algorithm.

\section{Conclusion}

In this work, we conduct a novel investigation into the potential and limitations of the widely used polynomial graph filter learning approach in graph modeling. Our preliminary study reveals the suboptimality and instability of the existing learning approaches, and we further uncover the severe overfitting issues as a plausible explanation for its failures. 
To address these limitations, we propose Auto-Polynomial, a novel and general automated polynomial graph filter learning framework to improve the effectiveness and generalization of any polynomial-based GNNs, making this work orthogonal and complementary to existing efforts in this research topic. Comprehensive experiments and ablation studies demonstrate the significant and consistent improvements of Auto-Polynomial in various learning settings. This work further unleashes the potential of polynomial graph filter-based graph modeling.

\bibliographystyle{ACM-Reference-Format}
\bibliography{ref}

\appendix

\end{document}